%% file: main.tex
\documentclass[runningheads]{llncs}
\usepackage[usenames,dvipsnames]{xcolor}
\definecolor{r0}{RGB}{103,0,31}
\definecolor{r1}{RGB}{178,24,43}
\definecolor{r2}{RGB}{214,96,77}
\definecolor{r3}{RGB}{244,165,130}
\definecolor{r4}{RGB}{253,219,199}
\definecolor{w0}{RGB}{247,247,247}
\definecolor{b4}{RGB}{209,229,240}
\definecolor{b3}{RGB}{146,197,222}
\definecolor{b2}{RGB}{67,147,195}
\definecolor{b1}{RGB}{33,102,172}
\definecolor{b0}{RGB}{5,48,97}

\definecolor{bl}{RGB}{166,206,227}
\definecolor{bd}{RGB}{31,120,180}
\definecolor{gl}{RGB}{178,223,138}
\definecolor{gd}{RGB}{51,160,44}
\definecolor{pinkp}{RGB}{251,154,153}
\definecolor{redp}{RGB}{227,26,28}
\definecolor{ol}{RGB}{253,191,111}
\definecolor{od}{RGB}{255,127,0}
\definecolor{pl}{RGB}{202,178,214}
\definecolor{pd}{RGB}{106,61,154}
\definecolor{yellowp}{RGB}{255,255,153}
\definecolor{brownp}{RGB}{177,89,40}

\usepackage[bookmarks=true, bookmarksopen = true]{hyperref}
\usepackage[most]{tcolorbox}
\usepackage{printlen}
\usepackage[T1]{fontenc}
\usepackage{booktabs}

\usepackage{babel}
\usepackage[font=small,labelfont=bf]{caption}

\usepackage{todonotes}
\usepackage{amsfonts}
\usepackage{amsmath}
\usepackage{mathtools}
\usepackage{xspace}
\usepackage[inline]{enumitem}
\newlist{choices}{enumerate*}{1}
\setlist[choices]{label=(\roman*)} %itemsep = 1.125in,
\usepackage{float}

\usepackage[capitalise]{cleveref}

\usepackage{algorithm2e}
\makeatletter
\renewcommand{\@algocf@capt@plain}{above}% formerly {bottom}
\makeatother

\usepackage{graphicx}
\input{commands}
\begin{document}
%
% \title{Weakly-Supervised Lesion Segmentation via Counterfactual Diffusion}
% \title{Normal to Abnormal: Lesion Localization via Counterfactual Diffusion with Implicit Guidance}
%\title{What is Healthy? Lesion Localization via Counterfactual Diffusion with Implicit Guidance}
\title{What is Healthy? Generative Counterfactual Diffusion for Lesion Localization}
%Could implicit generation of counterfactual localize lesions?%
%\title{Guiding Brain Images from Healthy to Unhealthy with Diffusion Models}
%\title{Implicitly Guiding Brain Images from Normal to Abnormal with Diffusion Models}
%\title{Unhealthy Counterfactual with Diffusion Models}
\titlerunning{Lesion Localization via Generative Counterfactual Diffusion}
% If the paper title is too long for the running head, you can set
% an abbreviated paper title here
%

\author{Pedro Sanchez\inst{1,2} \and Antanas Kascenas\inst{2,3} \and Xiao Liu\inst{1, 2} \and Alison Q. O'Neil\inst{2,1} \and \\ Sotirios A. Tsaftaris\inst{1,2,4}}
\authorrunning{Sanchez et al.}
\institute{The University of Edinburgh, Edinburgh, Scotland, UK \and Canon Medical Research Europe, Edinburgh, Scotland, UK \and University of Glasgow, Glasgow, Scotland, UK \and The Alan Turing Institute, London, UK}
\maketitle              % typeset the header of the contribution
\sloppy

\begin{abstract}

Reducing the requirement for densely annotated masks in medical image segmentation is important due to cost constraints. In this paper, we consider the problem of inferring pixel-level predictions of brain lesions by only using image-level labels for training. By leveraging recent advances in generative diffusion probabilistic models (DPM), we synthesize counterfactuals of ``How would a patient appear if $X$ pathology was not present?''. The difference image between the observed patient state and the healthy counterfactual can be used for inferring the location of pathology. We generate counterfactuals that correspond to the minimal change of the input such that it is transformed to healthy domain. This requires training with healthy and unhealthy data in DPMs. We improve on previous counterfactual DPMs by manipulating the generation process with implicit guidance along with attention conditioning instead of using classifiers.\footnote{Code is available at \url{https://github.com/vios-s/Diff-SCM}.}

\end{abstract}

\section{Introduction}

Despite being crucial for training supervised machine learning models, pixel-level annotations of pathologies are costly. Creation of ground truth masks requires specialist radiologists. In this paper, we explore how to reduce the need for densely annotated ground truths in favour of a single image-level label: ``Is the patient healthy or not?''.

This problem has been tackled in the anomaly segmentation literature \cite{vae1,vae2,vaegradient,restoration,anogan,pinaya2022fast,kascenas2022denoising} by training deep generative models on healthy data only. They rely on the assumption that the model will learn ``normal'' (i.e.~ healthy) features whilst failing on out-of-distribution features~\cite{comparative}. In this case, a pixel-wise map can be generated by taking the residual between the input image and a predicted ``healthy'' image, to highlight the unhealthy areas. However, distinguishing normal (healthy) from abnormal (unhealthy) without being shown examples of abnormality is not trivial. In a brain lesion segmentation task~\cite{brats3}, for instance, lesions deform adjacent areas of the brain. Arguably, these deformations should not be captured by the anomaly segmentation algorithm. In fact, it has been hypothesised that anomaly segmentation models \cite{comparative} (trained only on healthy data) simply highlight zones of hyper-intensity in the image, since they may be surpassed in segmentation performance by simple thresholding techniques \cite{thresholding}. In line with this hypothesis, we show that, despite being more expressive than previous generative models,~\footnote{such as variational autoencoders (VAEs), normalizing flows (NFs) or generative adversarial networks (GANs)} diffusion probabilistic models (DPM)~\cite{Ho2020DenoisingModels,dhariwal2021diffusion} trained on healthy data \textit{only}, do \textit{not} perform well in the segmentation task.

%We show that similar anomaly segmentation strategies using diffusion probabilistic models (DPM)~\cite{Ho2020DenoisingModels} trained on healthy \textbf{only} do \textbf{not} perform well in the segmentation task albeit being more expressive than previous generative models~\cite{dhariwal2021diffusion}. DPMs are a family of generative models proven~\cite{dhariwal2021diffusion} to be more expressive than variational autoencoders (VAEs) or generative adversarial networks (GANs). 
DPMs synthesise images by decomposing the generation process into a sequential application of denoising neural networks. We show empirically that efficient localization of brain lesions (abnormalities) with DPMs requires showing the model during training what an unhealthy brain is. Here, we demonstrate that the areas of interest (brain lesions) will be highlighted by performing the \textit{minimal} intervention that can be applied to an image in order to change domains. This can be done using counterfactual generation~\cite{Pawlowski2020DeepInference,sanchez2022diffusion}. In particular, a recent technique for generation of image counterfactuals~\cite{sanchez2022diffusion} leverages a classifier (trained on image-level information) for manipulating an image between domains without paired domain data. The image is manipulated by encoding to latent space followed by conditional decoding. We create heatmaps by taking the difference between the observed image of a patient and its healthy counterfactual. However, using an extra classifier~\cite{sanchez2022diffusion,wolleb2022diffusion} can be cumbersome and hard to tune due to gradient dynamics during the iterative inference process of DPMs. In this paper, we improve on previous works by performing counterfactual diffusion without relying on downstream classifiers. We formulate a more efficient algorithm by using implicit guidance, attention-based conditioning and dynamic normalisation inspired by text-to-image DPMs~\cite{saharia2022photorealistic,Rombach2022Latent}.

\noindent\textbf{Contributions.}~We explore brain lesion segmentation with generative DPMs without pixel-level supervision. We:
\begin{choices}
    \item show that training DPMs on healthy data alone might not be sufficient for segmenting lesions, validating a previous hypothesis \cite{thresholding} that most anomaly segmentation algorithms only detect hyper-intensities;
    \item perform counterfactual diffusion \textit{without} relying on a downstream classifier, simplifying training and making the algorithm more robust to hyperparameter choices; and
    \item conduct extensive experiments, showing superior accuracy in brain lesion localization.
\end{choices}
%To the best of our knowledge, we are the first work to use DPMs for weakly-supervised segmentation without classifier guidance while achieving state-of-the-art results. Our method improves and simplifies conditioning with DPMs for counterfactual estimation by utilizing:

\begin{figure}[h]
\centering
\includegraphics[width=\textwidth]{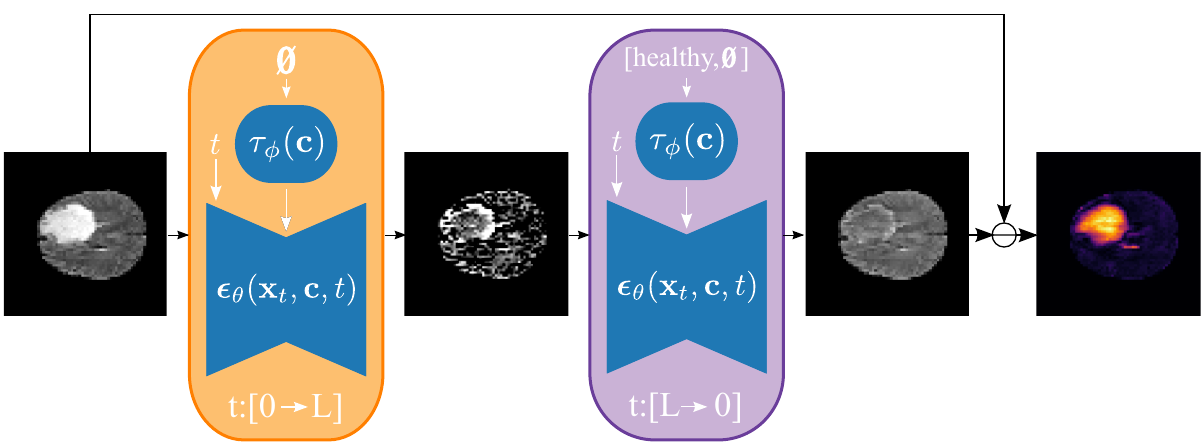}
\caption{Counterfactual diffusion overview. %The {\color{b1} \textbf{training}} consists of learning to predict the noise (with $t$-dependent variance) injected into an image. 
{\color{od}\textbf{Encoding}} is done by iteratively applying diffusion models to obtain a latent space with an unconditional model ($\bc = \emptyset$). {\color{pd}\textbf{Decoding}} is performed by reversing the diffusion process from the latent space to reconstruct an image with \textit{healthy} condition. As detailed in Sec.~\ref{sec:implicit_guidance}, decoding uses \textit{healthy} and $\emptyset$ for guidance. A heatmap of the lesion can obtained by taking the difference between the original and the reconstructed healthy.}
\label{fig:overview}
\end{figure}

\section{Preliminaries on Diffusion Probabilistic Models (DPMs)}

A diffusion process gradually adds noise to a data distribution over time. Diffusion probabilistic models (DPMs)~\cite{Ho2020DenoisingModels} learn to reverse the noising process, going from noise towards the data distribution. DPMs can, therefore, be used as a generative model.
The diffusion process gradually adds Gaussian noise, with a time-dependent variance $\alpha_t$, to a data point $\mathbf{x}_0 \sim p_{\text{data}}(\mathbf{x})$ sampled from the data distribution. Thus, the noisy variable $\mathbf{x}_t$, with $t \in \left[ 0,T\right]$, is learned to correspond to versions of $\mathbf{x}_0$ perturbed by Gaussian noise following %$p\left(\mathbf{x}_t \mid \mathbf{x}(t-1) \right)=\mathcal{N}\left(\mathbf{x}_t ; \sqrt{1-\beta_{t}} ~ \mathbf{x}(t-1), \beta_{t} \mathrm{I}\right)$. therefore
$p \left(\mathbf{x}_t \mid \mathbf{x}_0\right)=\mathcal{N}\left(\mathbf{x}_t ; \sqrt{\alpha_{t}} \mathbf{x}_0,\left(1-\alpha_{t}\right) \mathrm{I}\right)$, where $\alpha_{t}:=\prod_{j=0}^{t}\left(1-\beta_{j}\right)$ and $\mathrm{I}$ is the identity matrix. 

%As such, $p(\mathbf{x}_t) =\int p_{\text{data}}(\mathbf{x}) p(\mathbf{x}_t \mid \mathbf{x}) \mathrm{d} \mathbf{x} $ should approximate the data distribution $p(\mathbf{x}_0) \approx p_{\text{data}}$ at time $t = 0$ and a zero centered Gaussian distribution at time $t = T$. Generative modelling is achieved by learning to reverse this process using a neural network $\boldsymbol{\epsilon}_{\theta}$ trained to denoise images at different scales $\beta_t$. The denoising model effectively learns the gradient of a log-likelihood w.r.t. the observed variable $\nabla_{\mathbf{x}} \log p(\mathbf{x})$ \cite{Hyvarinen2005EstimationMatching}.

%\subsection{Conditioning}

%DPMs are powerful tools for conditioning the generation of images \cite{dhariwal2021diffusion,Rombach2022Latent,nichol2021glide,saharia2022photorealistic} in the pixel space -- editing~\cite{meng2022sdedit}, inpainting~\cite{saharia2021palette}, colorization~\cite{saharia2021palette}, super-resolution~\cite{saharia2021image}--, using texts prompts~\cite{ramesh2022hierarchical,saharia2022photorealistic} or label information \cite{dhariwal2021diffusion}. 
\subsection{Training}

With sufficient data and model capacity, the following training procedure ensures that the optimal solution to $\nabla_{\mathbf{x}} \log p_t(\mathbf{x})$ can be found by training $\boldsymbol{\epsilon}_{\theta}$ to approximate $\nabla_{\mathbf{x}} \log p_t(\mathbf{x}_t \mid \mathbf{x}_0)$\cite{Hyvarinen2005EstimationMatching}. The diffusion model can be implemented with a conditional denoising U-Net $\diffusionmodel(\bx_t, \bc, t)$ which allows controlling the synthesis process through inputs $\bc$. The training procedure is done learning a $\theta^*$ such that
\begin{equation}
\label{eq:trainingDDPM}
\theta^* = \underset{\theta}{\arg \min } ~ \E_{\bx_0, t, \epsilon} \left[ \left\| \diffusionmodel(\bx_t, \bc, t)- \epsilon \right\|_{2}^{2}\right],
\end{equation}
where $\bx_t = \sqrt{\alpha_t}\bx_0 + \sqrt{1 - \alpha_t} \epsilon$, with $\mathbf{x}_0 \sim p_{\text{data}}$ being a sample from the (training) data distribution, $t \sim \mathcal{U}\left(0, T\right)$ and $\epsilon \sim \mathcal{N}\left(0, \mathrm{I}\right)$ is the noise.

\subsection{Inference}
%\textbf{Inference.}~
Once the model $\boldsymbol{\epsilon}_{\theta}$ is learned using Eq.\ \ref{eq:trainingDDPM}, generating samples consists in starting from $\bx_T \sim \mathcal{N}(\mathbf{0}, \mathrm{I})$ and iteratively sampling from the reverse process with the diffusion model. Here, we will use the sampling procedure from Denoising Diffusion Implicit Models (DDIM, \cite{Song2021DenoisingModels}) which formulates a deterministic mapping between latents to images following
\begin{equation}
\label{eq:ddim_sampling}
\bx_{t-1} = \sqrt{\alpha_{t-1}} \underbrace{\left( \frac{\bx_t - \sqrt{1 - \alpha_t} \cdot \diffusionmodel(\bx_t, \bc ,t)}{\sqrt{\alpha_t}}  \right)}_{\hat{\bx}_0}  + \sqrt{ 1 - \alpha_{t-1}} ~ \diffusionmodel(\bx_t, \bc ,t).
\end{equation}
The DDIM formulation has two main advantages:
\begin{choices}
    \item it allows a near-invertible mapping between $\bx_T$ and $\bx_0$; and
    \item it allows efficient sampling with fewer iterations even when trained with the same diffusion discretization. This is obtained by choosing different under-sampling $t$ in the $[0,T]$ interval.
\end{choices}

\section{Lesion Localization with Counterfactual Diffusion}
\label{sec:main_methods}
We are interested in manipulating the input image from unhealthy\footnote{If the input is healthy, applying an intervention should not modify it.} to healthy domain at inference time. At the same time, all other aspects of the input image should remain unchanged. Specifically, we are interested in identifying what is the main feature that should be modified. The main features should, for instance, correspond to lesions in a brain tumour dataset. This ``minimal'' manipulation is known in the causal literature as counterfactual generation~\cite{Pawlowski2020DeepInference,sanchez2022diffusion}. Once $\diffusionmodel(\bx_t, \bc, t)$ is trained on the imaging data with $\bc \in [\text{healthy}, \text{unhealthy}]$ using Eq.~\ref{eq:trainingDDPM}, we can manipulate the input image between domains at inference time using the counterfactual generation inspired by \cite{sanchez2022diffusion,wolleb2022diffusion}.

%DPMs can be used to manipulate images, creating counterfactuals . Here, we use the conditional synthesis capabilities of DPMs for generating healthy counterfactual images of MRI images. We first detail some of the techniques that allow us to effectively use conditioning information for counterfactual generation such as attention conditioning, dynamic normalization and classifier-free guidance. We then proceed to define an algorithm to extract brain lesion heatmaps that can be thresholded to obtain segmentations. The overall methodology is illustrated in Fig.\ \ref{fig:overview}.

\subsection{Estimating Lesion Heatmap with Counterfactual Diffusion}
\label{sec:counterfactual_diffusion}
We {\color{od} \textbf{encode}} the input image into a (spatial) latent space by iteratively ($L$ iterations) applying Eq.\ \ref{eq:ddim_sampling} in reverse order (simply changing $t-1$ to $t+1$) with an unconditional model ($\bc = \emptyset$). Then, we generate a counterfactual by {\color{pd} \textbf{decoding}} the latent while applying an intervention to the conditioning $\bc$ to be "healthy". Decoding is done by applying Eq.\ \ref{eq:ddim_sampling} with implicit guidance (using $\diffusionmodel(\bx_t, \bc, t)$ as in Sec.~\ref{sec:implicit_guidance}) with attention conditioning (Sec.~\ref{sec:attention_conditioning}). The difference between the original image and counterfactual is then averaged along the channel dimension to obtain a heatmap which is used to recover the lesion (unhealthy features) segmentation. We apply dynamic normalization (Sec.~\ref{sec:dynamic_normalization}) throughout the entire inference process. We illustrate this method on Fig.\ \ref{fig:overview} and detail the algorithm in Alg.\ \ref{alg:counterfactual_estimation}.

\subsection{Implicit Guidance}
\label{sec:implicit_guidance}
Using a classifier \cite{dhariwal2021diffusion} to guide the diffusion process, which requires training an extra model over noisy images, has been successfully used to generate counterfactuals \cite{sanchez2022diffusion,wolleb2022diffusion}. 
%However, classifier guidance combines the gradients of a classifier with output of diffusion process. In addition, the classifier needs to be trained over noisy images and be conditioned on $t$, therefore, one cannot simply plug a standard pre-trained classifier. Classifier guidance requires training of an extra model. 
Here, we leverage \emph{implicit guidance}\footnote{Also known as \textit{classifier-free} guidance in text-to-image generation DPMs~\cite{ramesh2022hierarchical,saharia2022photorealistic}.}~\cite{Ho2021classifierfree} in the context of counterfactual generation. In implicit guidance, a single diffusion model is trained on conditional and unconditional objectives via randomly dropping~$\bc$ during training (e.g. with 35\% probability). The dropped conditioning is represented here with $\emptyset$ such that $\diffusionmodel(\bx_t, \bc, t)$ and $\diffusionmodel(\bx_t, \emptyset, t)$ are conditional and unconditional $\diffusionmodel$-predictions. Sampling is performed by combining $\diffusionmodel$-predictions with a guidance \emph{scale} ($w$), resulting in $\diffusionmodel(\bx_t, \bc, t) = w\diffusionmodel(\bx_t, \bc, t) + (1-w)\diffusionmodel(\bx_t, \emptyset, t)$. %Setting $w = 1$ disables classifier-free guidance, while increasing $w > 1$ strengthens the effect of guidance.

\subsection{Conditioning with Attention Mechanisms}
\label{sec:attention_conditioning}
Generating counterfactuals requires conditioning the decoding during inference. As baseline, we utilise the adaptive group normalization (AdaGroup) which has already been successfully used in DPMs \cite{dhariwal2021diffusion}. For counterfactual generation, modifying the normalization is not enough. We improve conditioning by augmenting the underlying U-Net backbone with a conditional attention mechanism inspired by previous work of text-to-image generation\cite{Rombach2022Latent,nichol2021glide,ramesh2022hierarchical}. To pre-process $\bc$, we use an encoder $\conditioner$ that projects $\bc$ to an intermediate representation $\conditioner(\bc) \in \R^{d_\tau \times d_\tau}$, which is separately projected to the dimensionality of each attention layer throughout the model, and then concatenated to the attention context at each layer. In particular, we consider a U-Net with an attention layer implementing $\text{softmax}\left(\frac{QK_\bc^T}{\sqrt{d}}\right) V_\bc$. Similar to previous DPMs~\cite{Ho2020DenoisingModels,dhariwal2021diffusion}, the values for $Q$, $K$ and $V$ are derived from the previous convolutional layer. However, we concatenate $\conditioner(\bc)$ to $K$ and $V$ before the attention layer forming $K_\bc = \text{concat}([K, \conditioner(\bc)])$ and $V_\bc = \text{concat}([V, \conditioner(\bc)])$~\cite{nichol2021glide}.

\subsection{Dynamic Normalization}
\label{sec:dynamic_normalization}
During inference, the iterative process with guidance might change the input image statistics. This is specially cumbersome to counterfactual estimation because the latent space pixel values saturate (high absolute values). A saturated latent generates lower quality reconstructions and less manipulable images. Most DPM methods \cite{dhariwal2021diffusion,Rombach2022Latent,nichol2021glide} clip the image to the $(-1,+1)$ range at each iteration. This \emph{static} normalization ensures that the image can be appropriately processed by the neural network\footnote{high absolute values of the neural network's input can result in unstable behaviour.} but also results in a saturated latent. 
We avoid saturation by normalizing ($dn$), at each sampling step, the intermediate image to a certain percentile absolute pixel value. We use $th = \text{max}(1, \text{percentile}(|\hat{x}_0|,s))$, where $s$ is the desired percentage. Then, $\hat{x}_0$ is clipped to the range $(-th,th)$ and divided by $th$. This dynamic normalization pushes saturated pixels (those near $-1$ and $+1$) inwards, thereby actively preventing pixels from saturation at each step.

\section{Experiments}

\subsection{Experimental Setup}

\noindent\textbf{Dataset.}~We evaluate the lesion localization performance on the surrogate task of brain tumor segmentation using data from the BraTS 2021 challenge \cite{brats3}. This data comprises magnetic resonance (MR) imaging from four sequences T1, post-contrast T1-weighted (T1Gd), T2-weighted (T2), and T2 Fluid Attenuated Inversion Recovery (FLAIR) for each patient. The data has already been co-registered, skull-stripped and interpolated to the same resolution. We use dataset splits with 938, 62 and 251 patients for training, validation, and testing. 

\noindent\textbf{Training.}~The dataset has pixel-level annotations for the lesions. During training, we consider axial slices with at least one tumour voxel to be ``unhealthy'', and  ``healthy'' otherwise. For the data input to the models, we concatenate all four modalities at the channel dimension for each patient. We normalize (rescale) the pixel values in each modality of each scan by dividing by the 99th percentile foreground voxel intensity. All slices are downsampled to a resolution of 64$\times$64 for training, but all evaluation is done at 128$\times$128 (1.62mm/pixel) for fair comparison with baselines.

\noindent\textbf{Benchmarks.}~We compare our model's performance against five generative methods, we use
\begin{choices}
\item a standard VAE \cite{vae1,vae2};
\item f-AnoGAN \cite{anogan};
\item VAE with iterative gradient-based restoration \cite{restoration};
\item denoising autoencoder (DAE) with coarse noise \cite{kascenas2022denoising};
\item counterfactual diffusion model with classifier guidance \cite{wolleb2022diffusion}\footnote{We train \cite{wolleb2022diffusion} at a different resolution than the original method for fair comparison. Therefore, we fine-tune their hyperparameters on a validation set for maximum performance as in Fig.\ \ref{fig:hyperparameters}.}.
\end{choices}
Finally, we apply the simple thresholding approach from \cite{thresholding}. We use the hyperparameters from the original works for the deep learning methods but tune manually where necessary to improve training stability and performance. We detail in Tab.~\ref{tab:segmentation_results} if a benchmark method use only healthy data or the entire dataset during training.

\noindent\textbf{Baseline.}~We use the denoising U-net $\diffusionmodel(\bx_t, \bc, t)$ from \cite{dhariwal2021diffusion} as diffusion model and perform encoding-decoding inference as described in Sec.~\ref{sec:counterfactual_diffusion} as baseline entitled counterfactual DPM (CDPM). Following previous anomaly localization works \cite{comparative,kascenas2022denoising}, we also train a model $\text{CDPM}_\text{healthy}$ on healthy data only and inference is performed by encoding and decoding with an \textit{unconditional} model $\diffusionmodel(\bx_t, t)$. During the reconstruction process lesions should not be reconstructed because they are out-of-distribution. 

\noindent\textbf{Evaluation.}~We evaluate the lesion localization performance of the methods with two metrics
\begin{choices}
\item area under the precision-recall curve (AUPRC) at the pixel level computed for the whole test set;
\item Dice score which measures the segmentation quality using the optimal threshold for binarization found by sweeping over possible values using the test ground truth. $\lceil$Dice$\rceil$ represents the upper bound for the Dice scores that would be obtainable in a more practical scenario.
\end{choices}

\subsection{Brain Lesion Localization}

\begin{figure}[h]
\centering
\includegraphics[width=\textwidth]{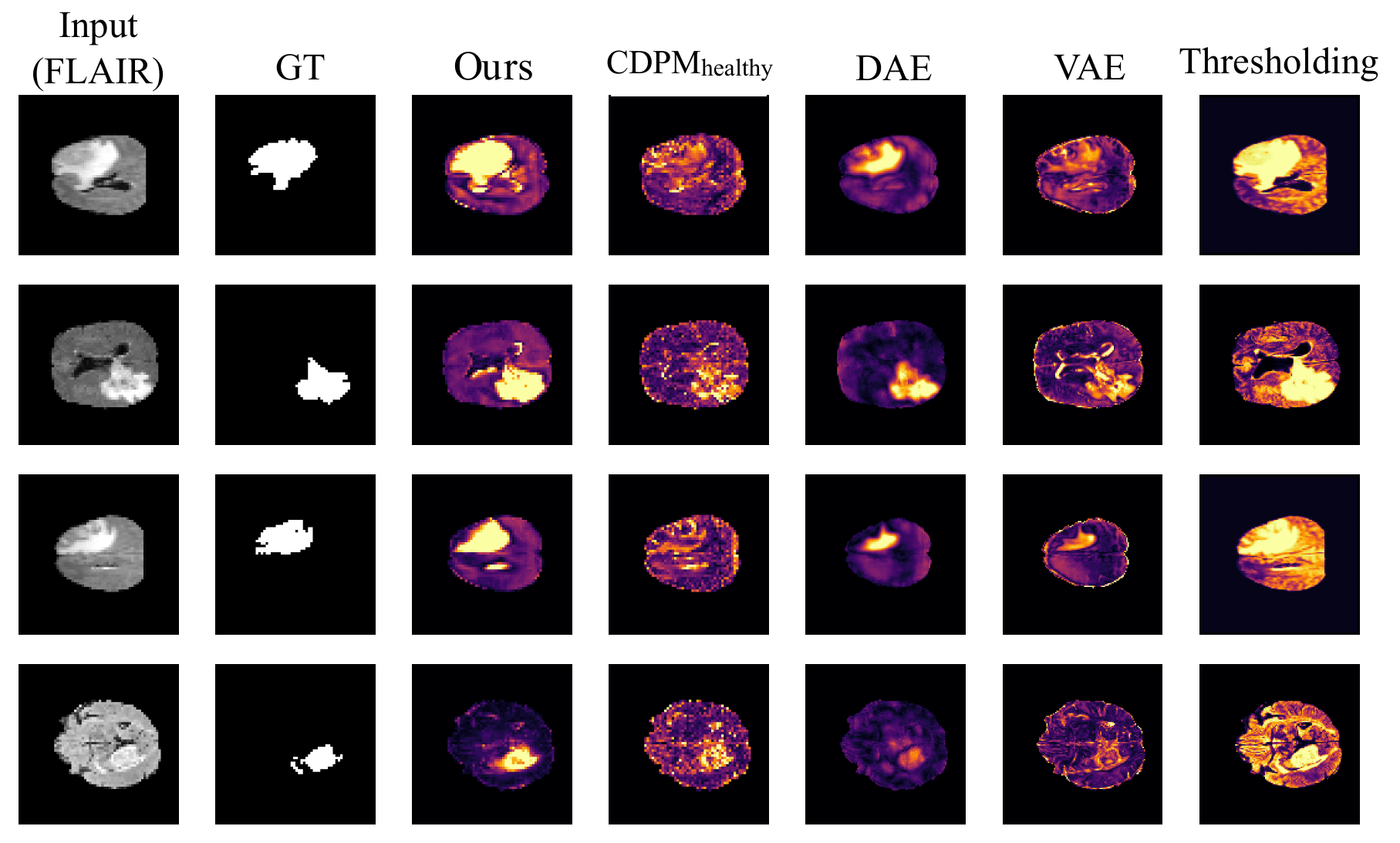}
\caption{Qualitative comparison of brain lesion segmentation methods. The columns indicate the input image (we are only illustrating the first channel but we use all four MRI sequences in our algorithm), the groud truth lesion masks and the heatmaps generated by each of the methods. Each row is a different slice.}
\label{fig:heatmap_comparison}
\end{figure}

We now compare our method to previous benchmarks on brain lesion localization. As shown in Tab.~\ref{tab:segmentation_results}, we surpass previous methods in a quantitative evaluation. In Fig.~\ref{fig:heatmap_comparison}, we show the qualitative difference between the heatmaps created by our  method when compared to other benchmarks. We also perform an ablation as shown in Tab.~\ref{tab:ablations}, studying the contribution of each individual components described in Sec.~\ref{sec:main_methods}.

\begin{table}[h]
\centering
\caption{Tumor detection performance as evaluated by test set wide pixel-level area under the precision-recall curve (AUPRC) and ideal Dice score ($\lceil$Dice$\rceil$). $\pm$ indicates standard deviation across 3 runs.}
\label{tab:segmentation_results}
\begin{tabular}{llll}
  \toprule
  \bfseries Method & \bfseries AUPRC & \bfseries $\lceil$Dice$\rceil$ & \bfseries Trained on \\
  \midrule
  Thresholding \cite{thresholding}  & 68.4 & 66.7 & Not \\
  %Thresholding + MF  & 0.798 & 0.749 & 0.750 \\
  f-AnoGAN~\cite{anogan} & 19.8{\scriptsize $\pm$0.6} & 31.6{\scriptsize $\pm$0.6} & Healthy \\
  VAE (reconstruction)~\cite{vae1,vae2}  & 29.9{\scriptsize $\pm$0.2} & 39.5{\scriptsize $\pm$0.2} & Healthy \\
  VAE (restoration)~\cite{restoration} & 74.0{\scriptsize $\pm$0.7} & 68.5{\scriptsize $\pm$0.5} & Healthy \\
  DAE~\cite{kascenas2022denoising}  & 81.6{\scriptsize $\pm$0.5} & 75.8{\scriptsize $\pm$0.4} & Healthy \\
  $\text{CDPM}_\text{healthy}$ & 24.9{\scriptsize $\pm$0.4} & 33.1{\scriptsize $\pm$0.4} & Healthy \\
  CDPM + classifier \cite{wolleb2022diffusion} & 81.5{\scriptsize $\pm$0.4} & 74.5{\scriptsize $\pm$0.4} & Full \\
  Ours & \textbf{82.8}{\scriptsize $\pm$0.4} & \textbf{76.2}{\scriptsize $\pm$0.3} & Full\\

  \bottomrule
  \end{tabular}
  
\end{table}

  \begin{minipage}{\textwidth}
  \begin{minipage}[h]{0.49\textwidth}
    \centering
    \includegraphics[width=\textwidth]{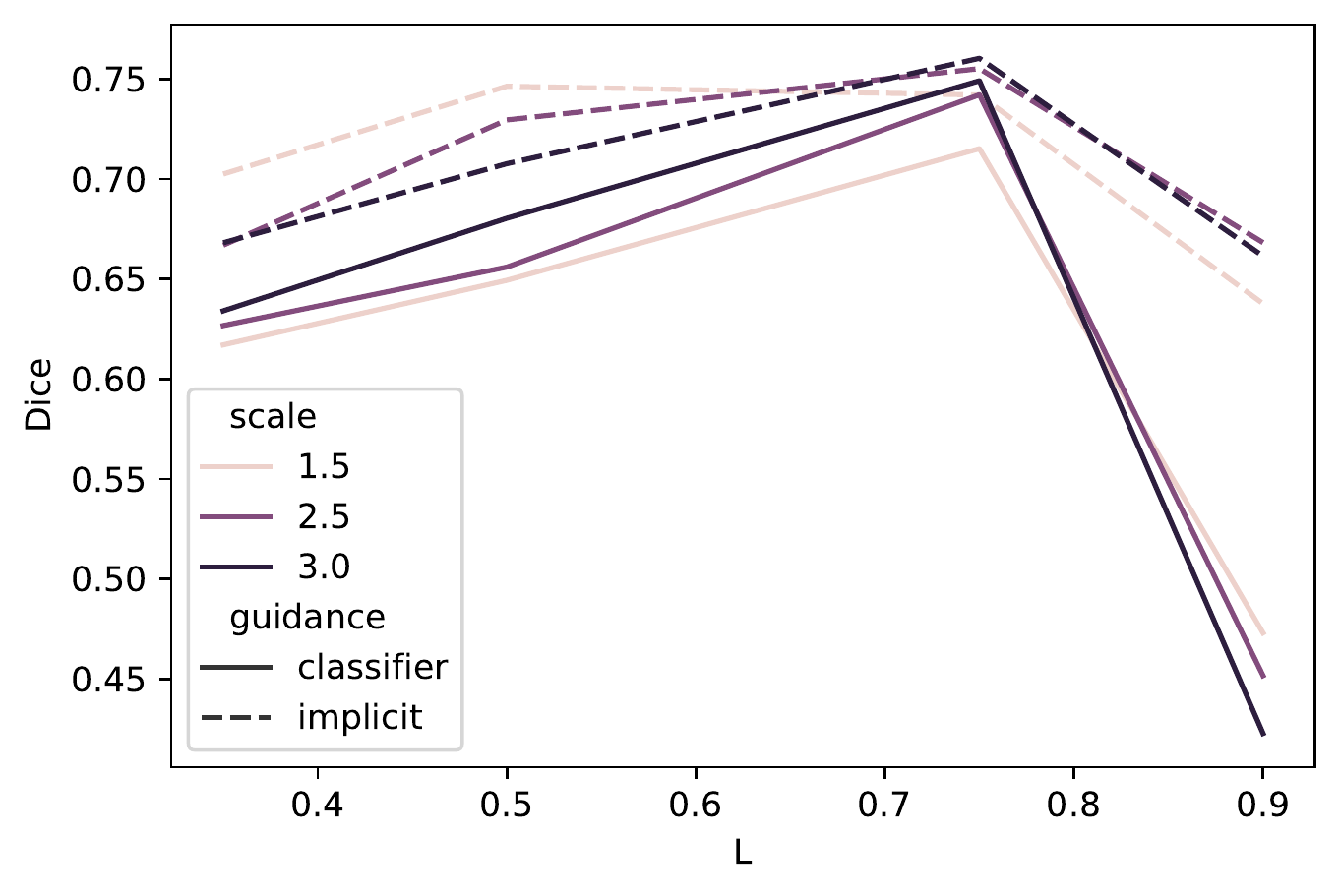}
    \captionof{figure}{Sweep through inference hyperparameters. We vary the guidance scale (both from classifier and implicit) and $L$ defined in terms of percentage of training diffusion steps (1 corresponds to 100 DDIM steps).}
    \label{fig:hyperparameters}
  \end{minipage}
  \hfill
  \begin{minipage}[h]{0.49\textwidth}
    \centering
    \captionof{table}{Contribution of each the components detailed throughout the paper to the localization results.}
    \label{tab:ablations}
    \begin{tabular}{l|l}
    \hline
    \multicolumn{1}{c|}{\textbf{Improvements}} & \multicolumn{1}{c}{\textbf{Dice}} \\ \hline
    
$\text{CDPM}_\text{healthy}$ & 33.1 \\
    \hline
    CDPM + Classifier \cite{wolleb2022diffusion} & 74.5 \\
    \hline
    CDPM  & 20.0 \\ 
    + implicit guidance           & 52.0 \\ 
    + attention conditioning             & 74.3 \\ 
    + dynamic normalisation             & \textbf{76.2} \\ 
    \end{tabular}

    \end{minipage}
  \end{minipage}

\subsection{How to Apply the Minimal Intervention?}

The brain tumours are the main feature differentiating healthy from unhealthy images. Therefore, we explore how to ensure that lesions are highlighted and nothing else, resulting in higher Dice values. In our algorithm, the strength of the intervention can be controlled by varying the guidance \emph{scale} ($w$) as well as the number of inference iterations $L$ at inference time. We use a small (256 images) annotated validation set to find the best inference hyperparameters by measuring Dice while varying other variables as illustrated in Fig.~\ref{fig:hyperparameters}. Our method (``implicit'') is more robust to hyperparameter choice than a counterfactual diffusion using classifiers \cite{wolleb2022diffusion}. 

%\subsection{Lesion Synthesis}

%We also explored using the model to generate unhealthy counterfactuals from healthy patients. Find qualitative results in Fig. \ref{fig:anomaly_synthesis}. {\color{red} this section might not be part of the final paper as we are not currently using the lesion synthesis for any downstream task.}
%\begin{figure}[t]
%\centering
%\includegraphics[width=0.7\textwidth]{figures/brain_anomaly_synthesis.png}
%\caption{Brain tumour synthesis. {\color{red} Details of this image will be improved. More samples?}}
%\label{fig:anomaly_synthesis}
%\end{figure}

\section{Related Works on Generative Models for Lesion Localization}

%\subsection{Non-supervised lesion segmentation}
Variational autoencoders (VAEs) \cite{vae1,vae2} constrain the latent bottleneck representation to follow a parameterized multivariate Gaussian distribution. \cite{vaegradient} further add a context-encoding task and combine reconstruction error with density-based scoring to obtain the anomaly scores, while \cite{restoration} use an iterative gradient descent restoration process at test time in restoration-VAE, replacing the reconstruction error with a restoration error to estimate anomaly scores. \cite{anogan} train a generative adversarial network called f-AnoGAN which reuses the generator and discriminator to train an autoencoder with both reconstruction and adversarial losses for the anomaly detection task. Recently, \cite{pinaya2022fast} combine a vector quantized VAE (VQ-VAE) to encode an image with a DPM model over the latent variables in order to produce reconstructions with fewer reproduced anomalies. Other works explored pseudo-healthy pathology synthesis by disentangling representations in generative adversarial networks (GANs) \cite{xia2020pseudo}.

\section{Conclusions}

We use conditional diffusion models for synthesizing healthy counterfactuals of a given input image, enabling lesion segmentation without access to pixel-level annotations. The difference between the observed image and the counterfactual produces a heatmap from which the segmentation masks can be obtained. Surprisingly, we show that this can be efficiently without a downstream classifier for guiding the generation, as in previous work \cite{wolleb2022diffusion,sanchez2022diffusion}. We show how using implicit guidance and attention conditioning as well as dynamic normalization, counterfactuals can be synthesized with a single model. Future work involves up-scaling the model to handle higher resolution images which can be done either by performing diffusion in a lower dimensional latent space \cite{Rombach2022Latent} or using a cascade of super-resolution conditional diffusion models \cite{saharia2022photorealistic}.

\section{Acknowledgements}

This work was supported by the University of Edinburgh, the Royal Academy of Engineering and Canon Medical Research Europe via PhD studentships of Pedro Sanchez and Xiao Liu (grant RCSRF1819\textbackslash825). This work was partially supported by the Alan Turing Institute under the EPSRC grant EP N510129\textbackslash1.

\bibliographystyle{splncs04}
\bibliography{ref}

\pagebreak

\appendix

\section{Algorithm}

\begin{algorithm}[h]
\caption{Segmentation with Implicit Counterfactual Diffusion} \label{alg:counterfactual_estimation}

\SetKwInOut{Model}{Model}
\SetKwInOut{Input}{Input}
\SetKwInOut{Hyperparameters}{Hyper-parameters~}
\SetKwInOut{Output}{Output}
\SetAlgoLined
\Model{trained diffusion model $\diffusionmodel$ }
\Hyperparameters{guidance scale $w$, number of iterations $L$}
\Input{factual image $\bx_{0}$ ($M$ channels), condition $\bc$}
\Output{heatmap}
%colback=pl!5!white,colframe=pl!50!black!50!
\begin{tcolorbox}[colback=pl!5!white,colframe=pd,colbacktitle=pl,left=2pt,right=2pt,top=0pt,bottom=1pt, title=\textbf{ \color{black} Recovering Unconditional Latent Space (Encoding)}] %
\For{$t \leftarrow 0$ \KwTo $L$}{
$\bxh_{t+1}  = dn \left( \sqrt{\alpha_{t+1}} \left( \frac{\bx_t - \sqrt{1 - \alpha_t} \cdot \diffusionmodel(\bx_t, \emptyset ,t) }{\sqrt{\alpha_t}}  \right) \right) + \sqrt{\alpha_{t+1}} \diffusionmodel(\bx_t, \emptyset ,t) $
}
\end{tcolorbox}

\begin{tcolorbox}[colback=ol!5!white,colframe=od,colbacktitle=ol,left=2pt,right=2pt,top=0pt,bottom=1pt,title=\textbf{\color{black} Counterfactual Generation (Decoding)}]
\For{$t \leftarrow L$ \KwTo $0$}{

$\epsilon = w~\diffusionmodel(\bx_t, \bc, t) + (1-w)~\diffusionmodel(\bx_t,\emptyset, t)$

$\bxh_{t-1}  = dn \left( \sqrt{\alpha_{t-1}} \left( \frac{\bx_t - \sqrt{1 - \alpha_t} \cdot \epsilon }{\sqrt{\alpha_t}}  \right) \right) + \sqrt{\alpha_{t-1}} \epsilon$
}
\end{tcolorbox}
heatmap = $\frac{1}{M}\sum^{M}_m |\bx_0^m - \bxh_0^m|$
\end{algorithm}

\end{document}

%% file: commands.tex
\newcommand{\E}{\mathbb{E}}
\newcommand{\R}{\mathbb{R}}

\newcommand{\bc}{\mathbf{c}}

\newcommand{\bx}{\mathbf{x}}

\newcommand{\bxh}{\hat{\mathbf{x}}}

\newcommand{\diffusionmodel}{{\boldsymbol{\epsilon}_{\theta}}}
\newcommand{\conditioner}{\tau_\phi}